\documentclass{bmvc2k}

\usepackage{multirow}
\usepackage{cleveref}
\usepackage{bbold}
\usepackage{amsmath,amssymb}

\usepackage{booktabs}
\usepackage{graphicx}

\usepackage{mathtools}

\usepackage[labelformat=simple]{subcaption}

\newcommand{\cameraready}[1]{#1}

\title{An Action Is Worth Multiple Words: \\ Handling Ambiguity in Action Recognition}

\addauthor{Kiyoon Kim}{kiyoon.kim@ed.ac.uk}{}
\addauthor{Davide Moltisanti}{davide.moltisanti@ed.ac.uk}{}
\addauthor{Oisin Mac Aodha}{oisin.macaodha@ed.ac.uk}{}
\addauthor{Laura Sevilla-Lara}{lsevilla@ed.ac.uk}{}

\addinstitution{
 School of Informatics\\
 University of Edinburgh \\
 Edinburgh, UK
}

\runninghead{Kim et al.}{Handling Ambiguity in Action Recognition}

\def\eg{\emph{e.g}\bmvaOneDot}

\def\ie{\emph{i.e}\bmvaOneDot}

\def\etc{\emph{etc}\bmvaOneDot}

\usepackage{fp}

\newcommand{\roundnumpm}[2]{\FPeval{\tempnum}{round(#1,1)}\tempnum $\pm$ \FPeval{\tempnum}{round(#2,1)}\tempnum}

\def\@fnsymbol#1{\ensuremath{\ifcase#1\or *\or \dagger\or \ddagger\or
   \mathsection\or \mathparagraph\or \|\or **\or \dagger\dagger
   \or \ddagger\ddagger \else\@ctrerr\fi}}
\newcommand{\nsymbol}[1]{\@fnsymbol{#1}}
\newcommand{\ssymbol}[1]{$^{\@fnsymbol{#1}}$}

\begin{document}

\maketitle

\begin{abstract}

Precisely naming the action depicted in a video can be a challenging and oftentimes ambiguous task. 
In contrast to object instances represented as nouns (e.g. dog, cat, chair, etc.), in the case of actions, human annotators typically lack a consensus as to what constitutes a specific action (e.g. jogging versus running). 
In practice, a given video can contain multiple valid positive annotations for the same action. 
As a result, video datasets often contain significant levels of label noise and overlap between the atomic action classes. 
In this work, we address the challenge of training multi-label action recognition models from only single positive training labels. 
We propose two approaches that are based on generating pseudo training examples sampled from similar instances within the train set. 
Unlike other approaches that use model-derived pseudo-labels, our pseudo-labels come from human annotations and are selected based on feature similarity. 
To validate our approaches, we create a new evaluation benchmark by manually annotating a subset of EPIC-Kitchens-100's validation set with multiple verb labels. 
We present results on this new test set along with additional results on a new version of HMDB-51, called Confusing-HMDB-102, where we outperform existing methods in both cases. 
Data and code are available at \href{https://github.com/kiyoon/verb\_ambiguity}{https://github.com/kiyoon/verb\_ambiguity}
\end{abstract}

\vspace{-15pt}
\section{Introduction}
\label{sec:intro}
\vspace{-5pt}

Actions are often fundamentally ambiguous. Barker and Wright \cite{barker1955midwest} explained that the hierarchical nature of an action is described differently depending on the viewpoint. In their example of ``children going to school", one could consider the atomic body movement (\eg ``going down the stairs") to the goal and intention of the action (\eg ``walking to the school", ``getting an education", \etc). This fundamental ambiguity and complexity in actions seems to influence the way humans learn verbs. %
For many languages, it has been observed that infants tend to learn nouns before verbs~\cite{waxman2013nouns}. 
Furthermore, according to WordNet~\cite{miller1995wordnet},
the number of unique nouns in the English vocabulary is more than ten times that of verbs, allowing for a more precise categorization. 
Similarly, researchers also face difficulties in the computational setting when naming, categorising and learning actions in videos
~\cite{wray2019learning}.

\begin{figure}[t]
    \centering
    \includegraphics[width=0.9\textwidth]{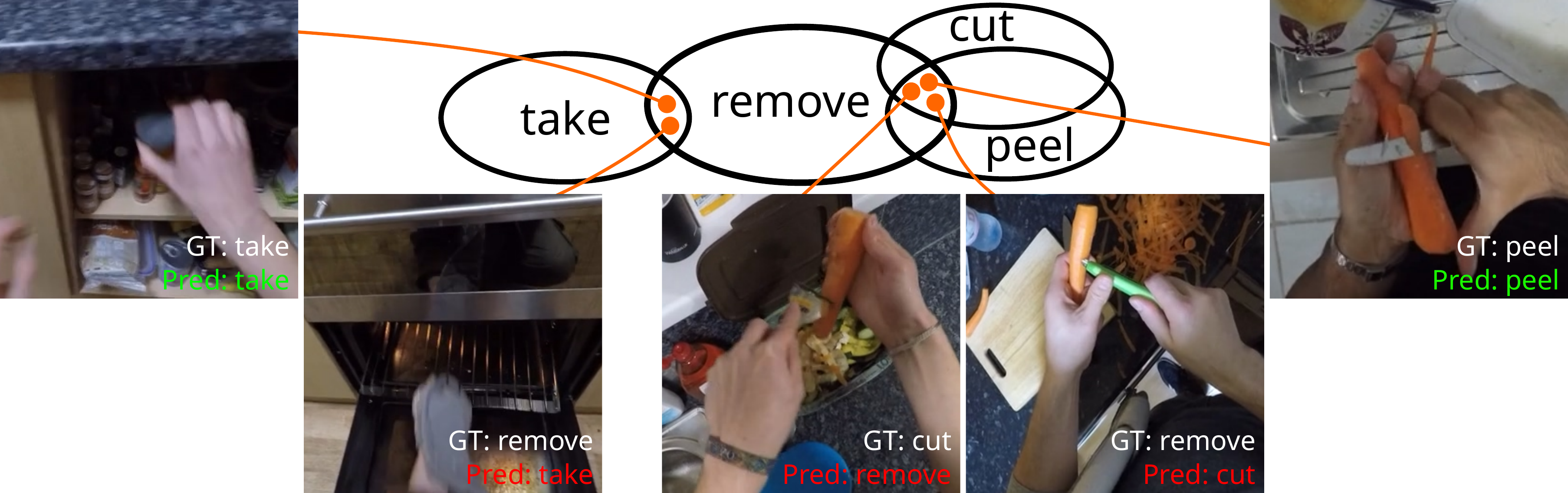}
    \caption{Ambiguity caused by single-verb labels. The same action can be described in different ways, \eg the act of peeling a carrot can be annotated as ``cut'', ``peel'' or ``remove'' (skin).
    Large action recognition datasets gather several instances with different but equivalent verb labels. Here we show some examples from the EPIC-Kitchens dataset~\cite{damen2018scaling}. Recognition models can be confused by this semantic ambiguity and are usually not evaluated taking this verb overlap into account. Reasonable and semantically correct predictions can be counted as incorrect (\eg red text, middle frames) by standard evaluation protocols as the predicted label does not match the noisy ground truth.}
    \label{fig:verb_ambiguity}
\end{figure}
Designing an action label space for a given dataset where each video is assigned a single unique label is an extremely difficult task. %
For example, consider the verbs ``stir'' and ``mix'' as labels for a cooking video. 
You can stir to mix the ingredients, but stirring is not the only way to mix them (\eg you can also ``shake''). Since each verb can often have multiple meanings, the label space can partially overlap. 
Specifically, one label can represent multiple types of actions, and many labels can represent a single type of action, as in \Cref{fig:verb_ambiguity}.
This is particularly difficult when datasets are large, and so is the label space, making it a combinatorial problem of finding overlapping classes for every instance. Labelling datasets in a thorough multi-label fashion can be prohibitively expensive. 
As a result, we typically only get partially-labelled video datasets, which negatively impacts both training and testing.

To address this, we propose a new training procedure to handle the problem of missing and ambiguous labels in action recognition. In particular, we generate pseudo-labels by searching neighbouring video instances in the feature space of the training set and analyse how similar videos are labelled differently. We take the commonly-agreed labels to be pseudo-labels (\ie labels that are highly likely to be positive and thus relevant for the query video) and train action recognition models to either not penalise the potentially-positive classes or boost the learning signal for them.
In order to evaluate our proposed methods we manually annotated a large subset of the validation set of the EPIC-Kitchens-100 dataset~\cite{damen2020rescaling} with multiple verbs.
We also evaluate our methods on a modified version of the popular HMDB-51 dataset \cite{kuehne2011hmdb}, where we added synthetic ambiguity. 
Overall, our methods improve multi-label accuracy up to \cameraready{18\%} over several recent baselines.
To summarise:
\begin{itemize}
    \vspace{-6pt}
    \item We show that the ambiguity in current action recognition labelling leads to issues in training and testing models. We show ``failure cases'' on challenging datasets where the model predicts reasonable labels but the accuracy is penalised because of label ambiguity.
    \vspace{-6pt}
    \item To mitigate the issue, we introduce two novel approaches for training  multi-label action recognition models when only single positive labels are available. Our approaches attempt to disambiguate the label space by generating pseudo-labels from similar instances that are labelled differently.
    \vspace{-6pt}
    \item We introduce a multi-label subset of the EPIC-Kitchens-100 validation set for the purpose of multi-label evaluation. On this dataset, and another with synthetically generated label ambiguity, we report results that outperform existing methods. 
\end{itemize}

\section{Related Work}

\noindent{\bf Action recognition.}
Action recognition refers to a video classification problem where the goal is to predict the action depicted in  a video. 
Many different approaches have been proposed. 
For example, 3D convolutions~\cite{c3d,i3d} have been widely used to capture spatio-temporal information, variants of which  include using relation networks~\cite{nonlocal} and capturing slow and fast motion in separate networks~\cite{slowfast}. 
A recent trend is to use transformer-based architectures~\cite{swin}. 
For efficiency, 2D networks have been a popular choice, and this includes averaging the predictions from 2D image classifiers~\cite{tsn}, using relation information~\cite{trn} or temporal shift modules~\cite{tsm} to share different frame features from the 2D backbone. 

\noindent{\bf Noisy labels and images.}
\cite{frenay2013classification} provides a summary of methods that learn from noisy labels.
Label smoothing~\cite{szegedy2016rethinking} is one popular and simple method to prevent models from being overly confident by converting hard labels to soft ones.
Progressive Self Label Correction~\cite{wang2021proselflc} is a label correction technique that mixes model predictions with labels. This method assigns a larger weight to the prediction of a given class if the model still outputs high confidence for a class after some initial training.
Other methods represent uncertainty with multiple hypotheses \cite{rupprecht2017learning}, use contrastive learning \cite{ortego2021multi}, correct the loss by estimating probabilities of one class being flipped to another~\cite{patrini2017making}, or optimise network parameters and labels alternatively~\cite{tanaka2018joint}.
These methods are commonly evaluated on image datasets with synthetic noise %
or on noisy image datasets collected from the web. 
In the latter case, noise derives from incorrect annotations or imperfect data collection. %
In our setting, noise stems from semantic ambiguity, \ie multiple equivalent labels can apply to the same action, which involves arguably %
more complicated dynamics compared to noise caused by labelling omissions.

\noindent{\bf Noisy labels and videos.}
Video datasets are generally noisy.~\cite{yeung2017learning} proposed a Q-learning method to mitigate the problem.~\cite{sharma2021noisyactions2m} introduced a video dataset where noise in the labels comes from inaccurate parsing of web tags. \cite{ghadiyaram2019large} explored pre-training methods using large-scale video datasets with label noise (\ie missing or incorrect labels) and temporal segmentation noise. 
\cite{Moltisanti_2017_ICCV} studied temporal ambiguity in action labelling showing that the perception of temporal bounds is highly subjective, which in turns leads to recognition robustness issues. 
Multiple instance learning has also been used to tackle noisy video datasets, where noise comes from less reliable annotations or sparse labelling \cite{leung2011handling,arnab2020uncertainty}.
However, none of these methods specifically address the noise generated by the ambiguous nature of verbs. %
This type of noise exhibits different distributions compared to noise stemming from the other problems discussed above. This is because the meaning of a verb varies depending on the video context, and as a consequence there is only a partial overlap across different labels, \ie labels are not exactly interchangeable across all videos.

\noindent{\bf Visual and semantic ambiguity.}
\cite{wray2016sembed} proposed a method to learn actions from visually similar instances with different semantic labels. Their framework builds a graph where nodes are connected if they are visually similar or semantically related. A Markov walk through this graph estimates the action label of a new video. This work is very close in spirit to ours, however experiments showed that it requires fine-grained verb meaning labels in order to achieve good results. %
On the contrary, we do not require further annotations and only use the available single-label ground truth. Furthermore, we release multi-label annotations for better testing of our methods.
\cite{wray2019learning} identified the issues arising from single-verb labels due to semantic ambiguity. They proposed training with multi-verb representations, which were obtained by asking annotators to label one representative sample per action with all verbs they deemed relevant. Verbs collected for a given action sample were then attached to the other instances of the action. This work is also closely related to ours, however we stress again that we only require %
single-label ground truth for training. %
We also show in our experiments that instance-based multi-verb representations better alleviate semantic ambiguity compared to global dataset-based representations.
Visual Sense Disambiguation~\cite{gella2016unsupervised} aims to identify the context of an action given an image and a verb. 
Understanding visual and semantic relationships in the label space has also been shown to improve multi-label image classification~\cite{deng2022beyond}.
\cameraready{~\cite{zhong2021polysemy} use language priors in the form of word embeddings to find relationships between verbs in  human-object interaction data. However, our preliminary experiments with word embeddings failed to produce sensible similarity estimates for the finer-grained datasets we use in this work.}
\cameraready{~\cite{goel2022not} address the problem of missing label imputation for scene graph generation proposing an iterative pseudo label approach}.

\noindent{\bf Single positive multi-label learning.}
There has been recent interest in the problem of multi-label learning from only single positive labels. The problem is outlined in~\cite{cole2021multi}, where the authors propose a loss that encourages models to predict both known and unlabelled positives. 
The loss also adds a regularisation term to match the expected number of positives per image.
\cite{zhou2022acknowledging} learns from the annotated positive labels using an entropy maximisation approach and decreases the loss penalty for unknown labels. 
Their method also treats samples with the lowest confidence as pseudo negatives. %
\cite{verelst2022spatial} outlined an image-specific spatial consistency loss that measures consistency in model predictions over multiple random crops from the input image.
A similar problem of multi-label learning from missing labels has been tackled by leveraging the estimated distribution of missing and known labels~\cite{zhang2021simple,durand2019learning}.

\noindent{\bf Semi-supervised learning.}
Semi-supervised learning approaches tackle the problem of having partially annotated labels, which translates to having noisy label estimates. This problem considers partially annotated instances in single-label classification tasks and does not entail multi-label classification.
In the context of image datasets, \cite{ghosh2021contrastive} showed that self-supervised contrastive learning can result in representations that are more robust to label noise. 
Contrastive learning has also been explored for action recognition by using only a small subset of the labels~\cite{singh2021semi}. 
Recently, an uncertainty-aware pseudo-label selection framework was proposed for both image and video~\cite{rizve2021defense}. 
Pseudo-labelling was also used to group potentially similar classes in~\cite{nassar2021all}. 
The major benefit of pseudo-labelling over contrastive learning is that it does not rely on domain-specific data augmentation.

\section{Problem and Methodology}
\label{sec:method}

As outlined above, in contrast to nouns, verb annotations are prone to semantic ambiguity. 
Such ambiguity stems from two issues: i) verbs have different meanings, thus a single verb can be attached to visually different actions and ii) individuals can often use different verbs to annotate the same action.
Figure~\ref{fig:verb_ambiguity} shows an example of semantic ambiguity, where the act of peeling a carrot in EPIC Kitchens~\cite{damen2018scaling} is labelled as ``peel'', ``cut'', and ``remove'' (skin). This is a common issue in large datasets where multiple annotators label videos and thus it is hard to enforce consistent classes. 
While noun labels alleviate semantic ambiguity, they add to the annotation burden and make learning actions a fragmented process. In fact, as noted in~\cite{wray2016sembed, wray2019learning}, nouns are mostly used to facilitate learning, however they unnaturally split the same action into different classes. 
For example, ``open-door'' and ``open-drawer'' are the same action, but belong to separate classes when categories are indicated by both verbs and nouns. Annotating actions with multiple verbs~\cite{wray2019learning} also mitigates ambiguity, however it comes with the cost of choosing multiple labels for each video.
Models are typically not designed, nor evaluated, taking this semantic ambiguity into account. 
We address the semantic ambiguity introduced by single-verb labels %
with two methods which generate multi-verb pseudo-labels from single-verb annotations during training. 
We also evaluate our methods across several metrics better suited for datasets with semantic ambiguity. We next formalise the problem and review common baselines, before introducing our methods.

\paragraph{Formalisation.}
We consider our problem to be related to that of  %
single positive multi-label learning (SPML)~\cite{cole2021multi,zhou2022acknowledging}. 
Here, we have a training and validation set where instances have only one positive label. Other categories are unknown and could be either positive or negative (\ie present or absent). 
For evaluation, instances in the test set have multiple labels. 
Formally, let $\Gamma = \{1, \dots, C \}$ be the set of classes in a given dataset, where $C$ is the number of classes. 
Each training/validation video is annotated with a single positive label $y_i \in \Gamma$. Remaining classes in $\Gamma \setminus \{y_i\}$ are not annotated and cannot be assumed to be negative.
A test video is annotated with a positive label set $\mathcal{Y}_i \subset \Gamma $ which can contain multiple classes, and remaining the classes $\Gamma \setminus \mathcal{Y}_i$ are treated as confirmed negatives.

\paragraph{SPML.} One of the most common approaches in SPML is the ``assume negative'' loss (AN)~\cite{cole2021multi}. This treats all unknown labels as negative ones and uses the binary cross-entropy (BCE) loss for multi-label learning. Given a training video-label pair ($\mathbf{x}_i$, $y_i$):

\begin{align}
    \mathcal{L_{\text{AN}}}(\mathbf{x}_i, y_i) = -\frac{1}{C} \sum_{c=1}^{C}{\left[ \mathbb{1}_{[y_i=c]}\log{(f^{(c)}(\mathbf{x}_i))} + \mathbb{1}_{[y_i\neq c]}\log{(1-f^{(c)}(\mathbf{x}_i))}\right]} 
\end{align}

\noindent where $\mathbb{1}_{[.]}$ is the indicator function and $f^{(c)}(\mathbf{x}_i)$ is the model prediction for class $c$.
To address the imbalance between the known positive labels and the unknown assumed-negative labels, an alternative version of AN is proposed: the ``weak assume negative'' loss (WAN)~\cite{cole2021multi}.
This improves over AN in that it gives equal weight for all assumed negatives and %
the single positive, however all unknown labels are still treated as negatives. Label smoothing (LS)~\cite{szegedy2016rethinking} is another popular approach to prevent models from being overly confident.
A variant of LS only for assumed negative labels (N-LS) was also proposed in \cite{zhou2022acknowledging}.
This improves over LS in that it only alters unknown (assumed negative) labels. The focal loss \cite{lin2017focal}, originally designed for object detection, has been used for SPML as well.
This works well in the presence of  imbalanced labels, however it requires additional parameter tuning to achieve optimal performance. 
Finally, the entropy maximisation (EM) loss \cite{zhou2022acknowledging} allows the unknown labels to be unknown rather than assuming them to be negatives. EM learns mainly from annotated labels and attains state-of-the-art results in SPML on \emph{image} datasets. We provide details for each of these losses in the supplementary material and compare our methods against these losses in Section~\ref{sec:results}.

\subsection{Mask and Pseudo+Single-label BCE}

\begin{figure}[t]
    \centering
    \resizebox{1.0\linewidth}{!}{
    \includegraphics[width=\textwidth]{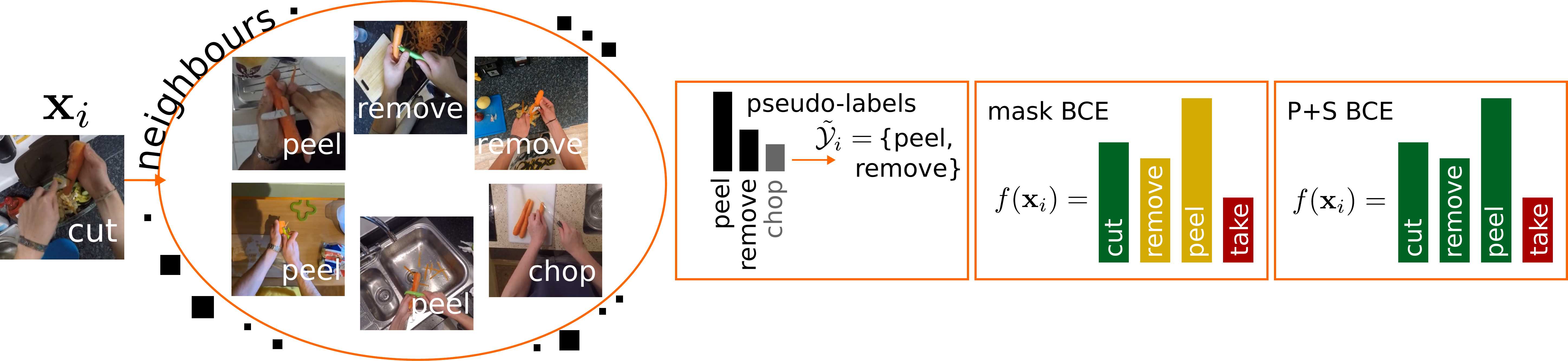}
    }
    \caption{Summary of our approach. Given a training video $\mathbf{x}_i$ we first retrieve its nearest neighbours in the feature space. Looking at the frequency of the neighbours' classes we then obtain the pseudo-labels set $\tilde{\mathcal{Y}_i}$. Pseudo-labels are used to optimise the model given its predictions $f(\mathbf{x}_i)$. The model is forced to predict the ground truth class (green bar) and penalised when predicting negative classes with high confidence (red bar). When using our Mask BCE loss, the model predictions for the pseudo-labels (yellow bars) are masked out during back-propagation, thus the model is neither penalised nor rewarded for guessing reasonable classes. When using our P+S BCE loss, pseudo-labels are assumed to be positive classes, thus the model is forced to predict both the ground truth and the pseudo-labels.}
    \label{fig:method}
\end{figure}

The above methods %
were mostly designed for partial labelled setting, where instances potentially contain multiple classes but \textit{classes do not semantically overlap}. In this setting there are multiple ``things'' present in an instance and the task is to label all of them, using only the available single-label annotations at training time. %
In our case, however, instances can belong to multiple classes \textit{because classes can semantically overlap}. %
We thus %
treat the unknown labels such that the model does not get penalised if it predicts an unknown label with high likelihood. 
Specifically, we pick classes that are likely to be positive based on feature similarity. We treat these as pseudo-labels. The model learns from single-verb ground truth annotations, however it is not penalised if it predicts a pseudo-label with confidence. %
Classes that are neither ground truth nor pseudo-labels are assumed negative.

To obtain pseudo-labels we extract features for each sample in the training set using a pre-trained action recognition backbone. 
Given a training pair $(\mathbf{x}_i, y_i)$ we take $K$ neighbouring videos in the feature space. These videos are likely to share the same label as $\mathbf{x}_i$, however we also expect these videos to belong to different classes given the semantically ambiguous setting we work in. 
We build pseudo-labels for $\mathbf{x}_i$ by looking at the label frequency of its $K$ neighbours. 
\cameraready{Formally, let $\Omega$ be the sequence of labels found among the $K$ neighbours, and let $\Upsilon$ be the set of unique labels in $\Omega$. For each $y_j \in \Upsilon$ we count its frequency in $\Omega$, \ie $\omega(y_j) = |( \eta \in \Omega : \eta = y_j)| / K$. The pseudo-label set $\tilde{\mathcal{Y}}_i$ is finally obtained from $\Upsilon$ by applying a threshold $\tau$ to $\omega(y_j)$, \ie  $\tilde{\mathcal{Y}}_i = \{y_j \in \Upsilon : \omega(y_j) > \tau, y_j \neq y_i\}$. 
}
Note that we do not add the original label to the pseudo-labels.
Unlike other methods~\cite{zhou2022acknowledging,wang2021proselflc} which generate pseudo-labels from model predictions, we use actual human annotations to produce pseudo-labels. 
With this approach, the ambiguous label space is realistically well-represented.
Clearly we rely on selecting informative neighbours, but we find empirically that the backbone we use is sufficient for this task.

With the above defined, we now propose our \textbf{Mask BCE} loss. 
This loss treats the single-label ground truth as the only positive, pseudo-labels as unknown classes, and assumes %
the remaining classes are negative:
\begin{align}
    \mathcal{L}_{\text{mask}}(\mathbf{x}_i, y_i) =
    -\frac{1}{C-|\tilde{\mathcal{Y}}_i|}
    \sum_{\substack{c=1\\ c \notin \tilde{\mathcal{Y}}_i}}^{C}{\bigg[ \mathbb{1}_{[y_i=c]} \log{\left( f^{(c)}(\mathbf{x}_i)\right) } + \mathbb{1}_{[y_i\neq c]} \log{\left( 1-f^{(c)}(\mathbf{x}_i) \right) }\bigg]} 
\end{align}
Here, the output of the model $f^{(c)}(\mathbf{x}_i)$ for classes $c \in \tilde{\mathcal{Y}}_i$ %
is detached from the gradient computation and back-propagation of the loss, \ie pseudo-label classes are frozen and do not provide any learning signal. 
This does not penalise the model if it produces reasonable predictions for relevant classes, \ie if it outputs a high likelihood for a class that is potentially an equivalent label for the video.
Importantly, this loss does not send incorrect positive learning signals when the pseudo-labels are not accurate. %
Instead, when pseudo-labels can be assumed to be correct, treating pseudo-labels as positive would be a better strategy.

With this in mind, we propose an alternative loss named \textbf{Pseudo+Single-label BCE}: %
\begin{align}
\begin{split}
    \mathcal{L}_{\text{P+S}}(\mathbf{x}_i, y_i) =
    -\frac{1}{C}
    \sum_{c=1}^{C}{\bigg[ \mathbb{1}_{[y_i=c]} \log{\left(f^{(c)}(\mathbf{x}_i)\right)}} &+ \mathbb{1}_{[c \in \tilde{\mathcal{Y}}_i]} \log{\left(f^{(c)}(\mathbf{x}_i)\right)} \\
    &+ \mathbb{1}_{[y_i \neq c]} \mathbb{1}_{[c \notin \tilde{\mathcal{Y}}_i]} \log{\left(1-f^{(c)}(\mathbf{x}_i) \right) \bigg] }
\end{split}
\end{align}
Intuitively, $\mathcal{L}_{\text{P+S}}$ will work better than  $\mathcal{L}_{\text{mask}}$ when the pseudo-labels are accurate, but it will hurt  performance when they are not. Figure~\ref{fig:method} summarises our pseudo-label generation and the two losses described above.

\section{Experiments}

\begin{figure}[t]
    \centering
    \includegraphics[width=\textwidth]{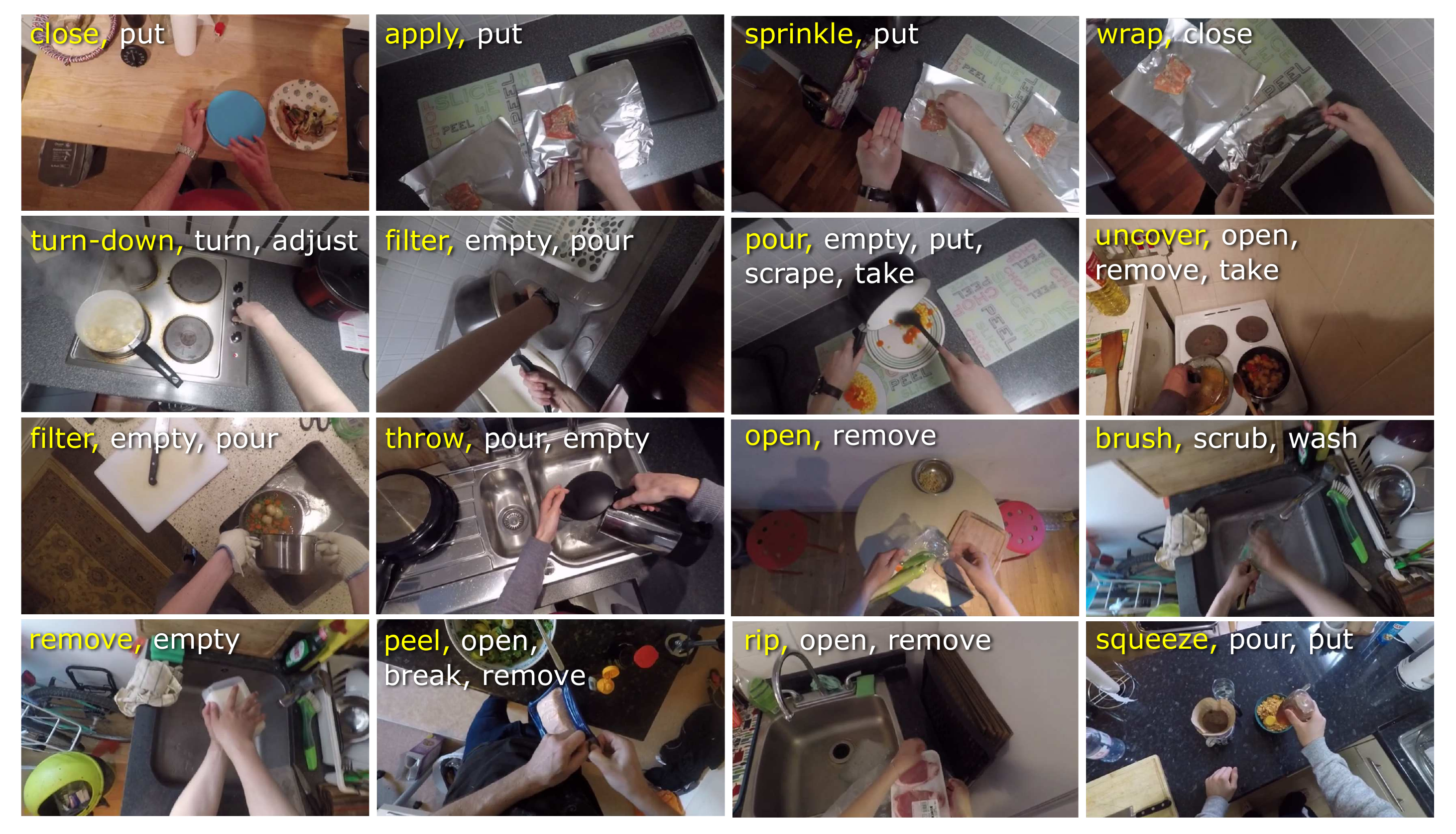}
    \vspace{-10pt}
    \caption{Samples from our EPIC-Kitchens-100-SPMV dataset. The yellow text indicates the original label from EPIC-Kitchens-100~\cite{damen2020rescaling}, which is also included in our annotations. The white text shows the additional labels we gathered from multiple annotators.}
    \label{fig:annotations}
\end{figure}

\begin{figure}[t]
    \centering
    \includegraphics[width=0.99\textwidth]{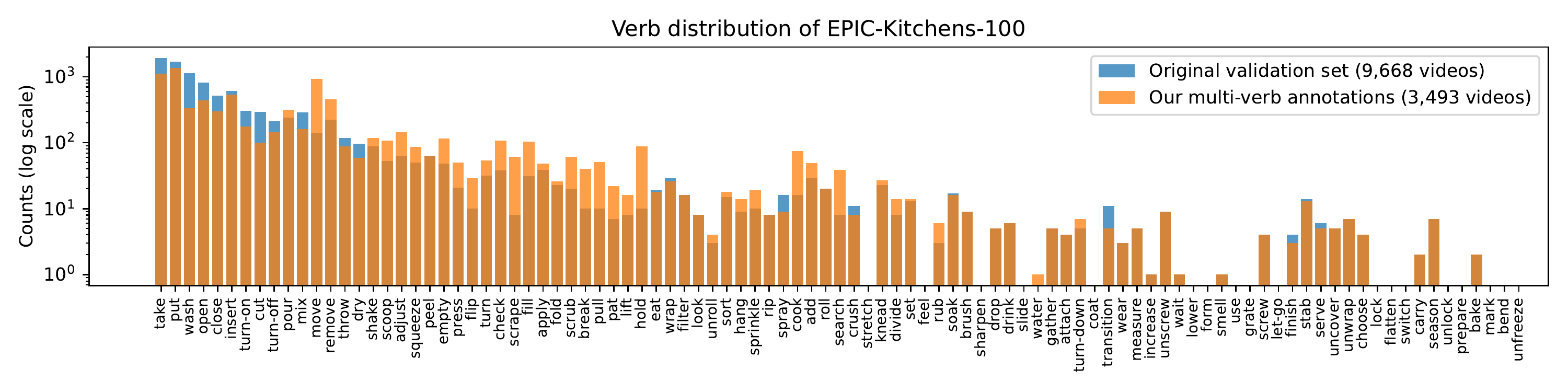}
    \vspace{-10pt}
    \caption{EPIC-Kitchens-100 verb distribution from our multi-label annotations (orange) compared to the original labels (blue). Although we only annotated 40\% of the validation dataset, we obtained similar or even more labels per class. Note that some generic verbs (\eg move, hold, search) gained a significant number of annotations compared to the original.}
    \label{fig:verb_distribution}
\end{figure}

\noindent{\bf Datasets.}
We manually annotated 40\% of the EPIC-Kitchens-100~\cite{damen2020rescaling} validation set with multiple verbs, using the procedure described in the supplementary material. 
This subset is our test set and consists of 3,493 multi-verb videos with 2.4 labels per video on average. The remaining %
videos in the original validation set form our own validation set. The training set %
is the same as the original one. \Cref{fig:annotations} illustrates a few examples from our dataset, while \Cref{fig:verb_distribution}
\cameraready{shows the verb distribution from our multi-label annotations and the original labels. Our annotations share a very similar distribution compared to the original ones, thus no bias towards specific classes is introduced with our labels}.
We call this dataset \textit{EPIC-Kitchens-100-SPMV} (Single Positive Multi-Verb learning). 
We also constructed a dataset from the popular HMDB-51 dataset \cite{kuehne2011hmdb} with synthetic labels in order to simulate ambiguous annotations.
We doubled the original 51 classes to 102 classes, where each class $c$ is split into $c^0$ and $c^1$. 
Given a training video $\mathbf{x}_i^{train}$ originally belonging to $c$, we randomly assign it to \textit{either}  $c^0$ or $c^1$. 
Given a test video $\mathbf{x}_i^{test}$ originally belonging to $c$, we assign it to  \emph{both} $c^0$ and $c^1$.
This simulates verb ambiguity, where actions are represented with multiple verbs. We name this dataset \textit{Confusing-HMDB-102}.

\noindent{\bf Metrics.}
We evaluate \textit{Top-Set Multi-label Accuracy} following~\cite{wray2019learning}.
This measures how many top predictions match the ground truth:
$A_{\text{Top-set\_ML}} = \frac{1}{N} \sum_{i=1}^N{|\mathcal{Y}_i \cap \hat{\mathcal{Y}}_i| / |\mathcal{Y}_i|}$, 
where $N$ is the total number of videos, $\mathcal{Y}_i$ is the multi-label ground truth, and $\hat{\mathcal{Y}}_i$ is the top-$K$ predicted classes, where $K$ is the number of ground-truth labels.
We also report another metric called \textit{Top-1 Multi-label Accuracy}:
$A_{\text{Top1\_ML}} = \frac{1}{N} \sum_{i=1}^N{\mathbb{1}_{[\hat{y}_i \in \mathcal{Y}_i]}}$,
where $\hat{y}_i$ is the top-1 predicted class.
This metric is more relaxed and does not penalise the model when it predicts a relevant verb, regardless of the specific choice. For example, when the ground-truth is (``stir'', ``mix''), the model is free to choose either ``stir'' or ``mix'', and not predicting the other one will not lower this accuracy. This is different from Top-Set Multi-label Accuracy, where predicting only one label in this case would give 50\% accuracy, and the model would have to predict both ``stir'' and ``mix'' to obtain 100\%.
We also evaluate \textit{IOU Accuracy}~\cite{sorower2010literature}, 
which is the intersection over union between the ground truth and predicted labels, 
and \textit{F\textsubscript{1}-Measure}, which is the harmonic mean of precision and recall. \cameraready{We set the confidence threshold to 0.5, so any predictions over 0.5 will be treated as positive labels.}
Finally, we report mean average precision \textit{mAP}. This is a popular metric for detection tasks. %
However, mAP is not very suitable for long-tail datasets such as EPIC-Kitchens, since class scores are averaged with equal weight.
\cameraready{We provide a formal definition of IOU Accuracy and F\textsubscript{1} in the supplementary material.}
For EPIC-Kitchens-100-SPMV, we choose the best model by looking at the standard top-1 accuracy on our single-label validation set, reporting results obtained on five different runs with different seeds.
For Confusing-HMDB-102 %
we take the best score per metric and take the average over the official three splits.

\noindent{\bf Implementation details.}
We pre-extracted RGB and optical flow features using TSM~\cite{tsm} pre-trained on each dataset's training set using a standard cross entropy loss with the single positive labels. %
TSM shares information across frames %
and has been shown to perform well on multiple action-orientated datasets. %
The concatenated RGB and optical flow features form the input to our model, which is a multi-layer perceptron (MLP) with three layers and hidden dimension of 1024. 
We train the MLP using the Adam optimiser~\cite{adamOptimiser} with a batch size of 64 and learning rate of 5e-6. We stopped training whenever the validation accuracy did not improve for 20 epochs. For EPIC-Kitchens-100-SPMV all experiments took less than 130 epochs to saturate. 
For Confusing-HMDB-102 it took longer, less than 390 epochs, given the much smaller size.
Pseudo-labels are generated only once before training. We set the number of neighbours $K=15$ and use a threshold of $\tau = 0.1$ to obtain pseudo-labels.

\subsection{Results}
\label{sec:results}

\Cref{tab:feature_results} compares our methods to multiple recent SPML baselines introduced in Section~\ref{sec:method}. 
We observe that our methods outperform all baselines across all metrics except mAP on EPIC-Kitchens-100-SPMV. 
As mentioned above, this metric is not well suited for imbalanced datasets.
We note that in most cases the P+S loss outperforms the Mask loss. 
Given that the P+S loss treats pseudo-labels as positives,
this demonstrates that our method for obtaining pseudo-labels is able to find good matches. 
We observe a clear boost %
especially in %
IOU  and F\textsubscript{1}-measure, which are well suited metrics for our multi-label setting. 
This indicates that the SPML baselines suffer from being penalised for correct (\ie related) classes. %
This also possibly suggests that SPML methods designed for partial-label settings struggle to work well when the source of the multi-label noise originates from semantic ambiguity. %

Interestingly, 
the EM~\cite{zhou2022acknowledging} loss, despite working well on static images, 
shows the worst %
IOU accuracy and F\textsubscript{1}-measure. We hypothesise this is because the loss %
does not treat unknown labels %
as negative.
This results in having no negative signal, which in turns gives overly confident predictions for most classes. %
\cameraready{Looking deeper into the model's output on EPIC-Kitchens-100-SPMV, the EM loss %
predicts 51.5 positive labels on average per video, 
whereas our mask method predicts on average 1.2 and P+S method %
predicts on average 1.8, with a confidence threshold of 0.5.} %
Considering that our test videos have %
2.4 positive labels on average, it is evident that the EM loss predicts \emph{too many} false positives. Our methods instead effectively push confidence up only for the relevant labels, keeping confidence for irrelevant labels low. %

\Cref{tab:ablation} (bottom) reports an ablation study on Confusing-HMDB-102 with an ideal label search, \ie the retrieved pseudo-labels are intentionally chosen as the correct classes. %
Results show that it is theoretically possible to improve performance with perfect pseudo-labels, however we do not observe a dramatic improvement compared to the labels obtained through our visual neighbour search.
\Cref{tab:ablation} (top) illustrates another ablation experiment, where we obtain pseudo-labels from class co-occurrences retrieved on our EPIC-Kitchens-100-SPMV test set. In this case, pseudo-labels are formed globally by looking at the co-occurrence of our multi-verb annotations across the test set. Given a verb $v_i$ we count how many times it was annotated together with any other verb $v_j$. 
If $v_i$ and $v_j$ occur at least half of the time together $v_j$ will form the pseudo-label set for instances labelled with $v_i$. %
The large performance drop observed when using these co-occurrence-based global pseudo-labels suggests that the semantic ambiguity in our setting requires \emph{instance-level} disambiguation rather than \emph{class-level} pseudo-labels. This is because verbs can have subtly different meaning which can change according to the context. Clustering verbs using co-occurrences ignores this signal, and thus hurts performance.  

\begin{table}[t]
\centering
\resizebox{0.90\linewidth}{!}{%
\begin{tabular}{@{}llccccc@{}}
\toprule
Dataset                                 & Loss       & Top-set ML         & Top-1 ML           & IOU Acc.           & F\textsubscript{1}                 & mAP                \\ \midrule
\multirow{8}{*}{EPIC-Kitchens-100-SPMV} & AN         & 43.7 ± 0.5          & 51.0 ± 0.4          & 11.2 ± 0.4          & 15.0 ± 0.6          & 22.8 ± 1.8          \\
                                        & WAN        & 44.8 ± 0.3          & 52.5 ± 0.6          & 15.2 ± 5.0          & 24.6 ± 6.6          & \textbf{26.3 ± 1.3} \\
                                        & LS         & 43.7 ± 0.9          & 51.8 ± 0.7          & 9.6 ± 0.4           & 12.9 ± 0.6          & 24.4 ± 1.5          \\
                                        & N-LS       & 43.4 ± 0.6          & 50.7 ± 0.4          & 11.0 ± 0.4          & 14.8 ± 0.5          & 22.1 ± 0.8          \\
                                        & Focal      & 43.3 ± 0.6          & 51.5 ± 0.3          & 5.7 ± 0.2           & 7.8 ± 0.2           & 23.9 ± 0.6          \\
                                        & EM         & 44.7 ± 0.4          & 52.9 ± 0.3          & 4.1 ± 0.0           & 7.8 ± 0.0           & 25.1 ± 1.0          \\ \cmidrule(l){2-7}
                                        & \cameraready{Mask (ours)} & \underline{46.6 ± 0.2}          & \underline{55.2 ± 0.4}          & \underline{27.8 ± 0.4}          & \underline{36.9 ± 0.4}          & \underline{25.9 ± 0.8}          \\
                                        & \cameraready{P+S (ours)}  & \textbf{46.9 ± 0.1} & \textbf{56.0 ± 0.6} & \textbf{33.5 ± 0.2} & \textbf{44.9 ± 0.3} & 25.8 ± 0.8          \\ \midrule 
\multirow{8}{*}{Confusing-HMDB-102}     & AN         & 32.0 ± 1.6          & 38.5 ± 2.0          & 18.9 ± 1.6          & 24.8 ± 1.6          & 20.6 ± 3.8          \\
                                        & WAN        & 36.8 ± 0.7          & 40.7 ± 0.6          & 4.1 ± 0.1           & 7.9 ± 0.1           & 32.0 ± 1.4          \\
                                        & LS         & 32.4 ± 2.3          & 38.8 ± 1.8          & 19.3 ± 2.3          & 24.9 ± 2.3          & 19.7 ± 3.8          \\
                                        & N-LS       & 32.2 ± 1.3          & 38.8 ± 1.7          & 19.3 ± 1.7          & 25.3 ± 1.8          & 20.1 ± 3.7          \\
                                        & Focal      & 31.6 ± 1.9          & 38.0 ± 2.0          & 13.0 ± 0.8          & 17.6 ± 0.7          & 14.8 ± 3.8          \\
                                        & EM         & 31.9 ± 0.6          & 37.4 ± 0.9          & 3.2 ± 0.0           & 6.2 ± 0.1           & 18.6 ± 3.1          \\ \cmidrule(l){2-7}
                                        & \cameraready{Mask (ours)} & \underline{41.8 ± 1.1}          & \underline{43.3 ± 0.9} & \textbf{30.8 ± 2.5}          & \textbf{36.3 ± 2.7}          & \underline{40.3 ± 2.2}          \\
                                        & \cameraready{P+S (ours)}  & \textbf{41.9 ± 0.9} & \textbf{43.4 ± 0.5}          & \underline{29.9 ± 2.2} & \underline{35.9 ± 2.0} & \textbf{40.7 ± 2.0} \\ \midrule
\end{tabular}%
}
\vspace{2pt}
\caption{Results with ± standard deviation calculated over five runs for EPIC-Kitchens-100-SPMV and three splits for Confusing-HMDB-102. For EPIC-Kitchens-100-SPMV the mAP metric shows unstable results due to the very long-tailed distribution of the verbs. Best and second-best results are bolded and underlined respectively.}
\vspace{-5pt}
\label{tab:feature_results}
\end{table}

\begin{table}[t]
\centering
\resizebox{0.90\linewidth}{!}{%
\begin{tabular}{@{}llccccc@{}}
\toprule
Dataset & Loss  & Top-set ML & Top-1 ML  & IOU Acc.  & F\textsubscript{1}    & mAP       \\ \midrule
\multirow{4}{*}{EPIC-Kitchens-100-SPMV} & \cameraready{Mask} & \underline{46.6 ± 0.2}          & \underline{55.2 ± 0.4}          & \underline{27.8 ± 0.4}          & \underline{36.9 ± 0.4}          & \textbf{25.9 ± 0.8}          \\
        & Mask\ssymbol{2}  & 37.8 ± 1.3  & 48.4 ± 1.0 & 18.1 ± 0.4 & 23.7 ± 0.4 & 21.7 ± 0.9 \\
        & \cameraready{P+S}  & \textbf{46.9 ± 0.1} & \textbf{56.0 ± 0.6} & \textbf{33.5 ± 0.2} & \textbf{44.9 ± 0.3} & \underline{25.8 ± 0.8} \\
        & P+S\ssymbol{2}   & 23.0 ± 0.4  & 28.0 ± 0.5 & 12.4 ± 0.4 & 18.0 ± 0.6 & 20.8 ± 1.3 \\ \midrule
\multirow{4}{*}{Confusing-HMDB-102}     & \cameraready{Mask} & 41.8 ± 1.1          & 43.3 ± 0.9 & \underline{30.8 ± 2.5}          & \textbf{36.3 ± 2.7}          & \underline{40.3 ± 2.2}          \\                    
        & Mask\ssymbol{1} & \underline{42.6 ± 2.2}  & \underline{43.8 ± 2.3} & 30.4 ± 2.3 & 34.2 ± 2.4 & 37.4 ± 3.6 \\
        & \cameraready{P+S}  & 41.9 ± 0.9 & 43.4 ± 0.5          & 29.9 ± 2.2 & \underline{35.9 ± 2.0} & \textbf{40.7 ± 2.0} \\
        & P+S\ssymbol{1}  & \textbf{43.2 ± 1.8}  & \textbf{44.3 ± 2.1} & \textbf{31.4 ± 2.5} & 35.9 ± 2.1 & 39.1 ± 3.1 \\ \midrule
\end{tabular}%
}
\vspace{2pt}
\caption{Ablation experiments using class-level pseudo labelling (\nsymbol{2}) or ideal pseudo-label search (\nsymbol{1}). We report results with ± standard deviation calculated over five runs for EPIC-Kitchens-100-SPMV and three splits for Confusing-HMDB-102. Best and second-best results are bolded and underlined.}
\vspace{-10pt}
\label{tab:ablation}
\end{table}

\section{Conclusion}
\vspace{-5pt}
We addressed the problem of action label ambiguity at training time in video action recognition. 
Here, videos can contain multiple equivalent labels due to the semantic ambiguity of verbs. %
This is a distinct problem compared to most partial-label settings, where multiple actions are present but not all are annotated.
We showed that state-of-the-art SPML approaches struggle in the presence of this semantic ambiguity.
To address this, we proposed two pseudo-labelled-based losses: \textit{Mask BCE}, which treats pseudo-labels neutrally %
and \textit{Pseudo+Single-label BCE}, which trusts pseudo-labels and uses them as positive labels. %
Through results on one dataset with synthetically generated ambiguity and another consisting of a new set of annotations for EPIC-Kitchens, we show that our methods outperform current SPML methods.  
One interesting direction for future research is addressing the additional ambiguity resulting from temporally overlapping actions that are often present in dense video datasets.

\small{
\par \noindent \textbf{Acknowledgements:} 
This work was in part supported by the Turing 2.0 `Enabling Advanced Autonomy' project funded by the EPSRC and the Alan Turing Institute. 
}

\pagebreak

\appendix

{\noindent\Large\textbf{\textcolor{bmv@sectioncolor}{Supplementary Material}}}

\section{EPIC-Kitchens-100-SPMV annotations}

\begin{figure}[t]
    \centering
    \includegraphics[width=0.7\textwidth,trim={0 4cm 0 4cm},clip]{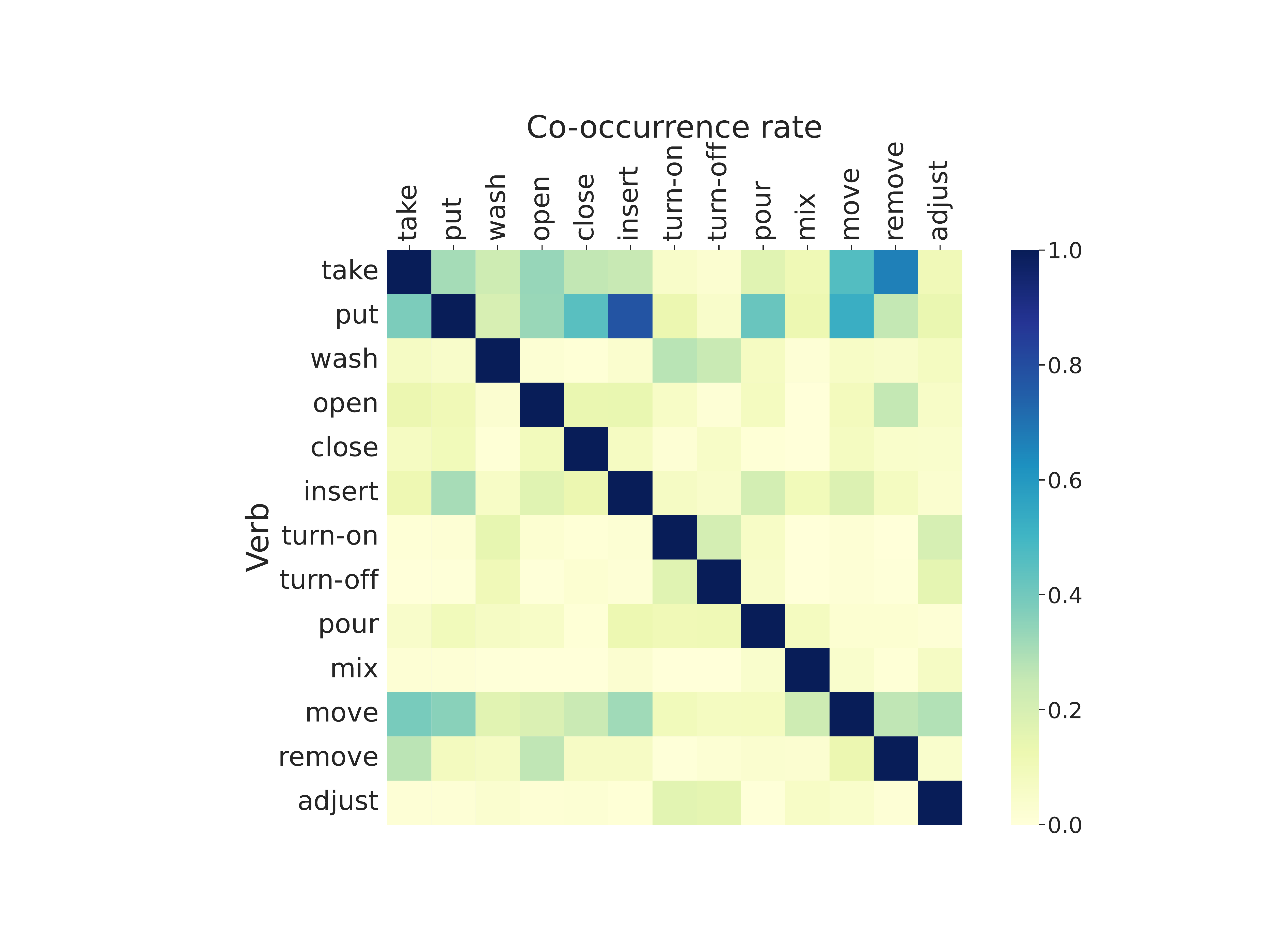}
    \vspace{-5pt}
    \caption{Co-occurrence ratio between verbs in our multi-label annotations.
    Rows are normalised by the number of classes, \eg 1 indicates two verbs always co-occur (diagonal), 0.5 means they occur half of the time together, etc. ``Take'' and ``remove'' frequently co-occur, and so do ``put'' and ``insert''. Only head classes are visualised.}

    \label{fig:multilabel_cooccurrence}
\end{figure}

\begin{figure}[t]
    \centering
    \includegraphics[width=0.7\textwidth]{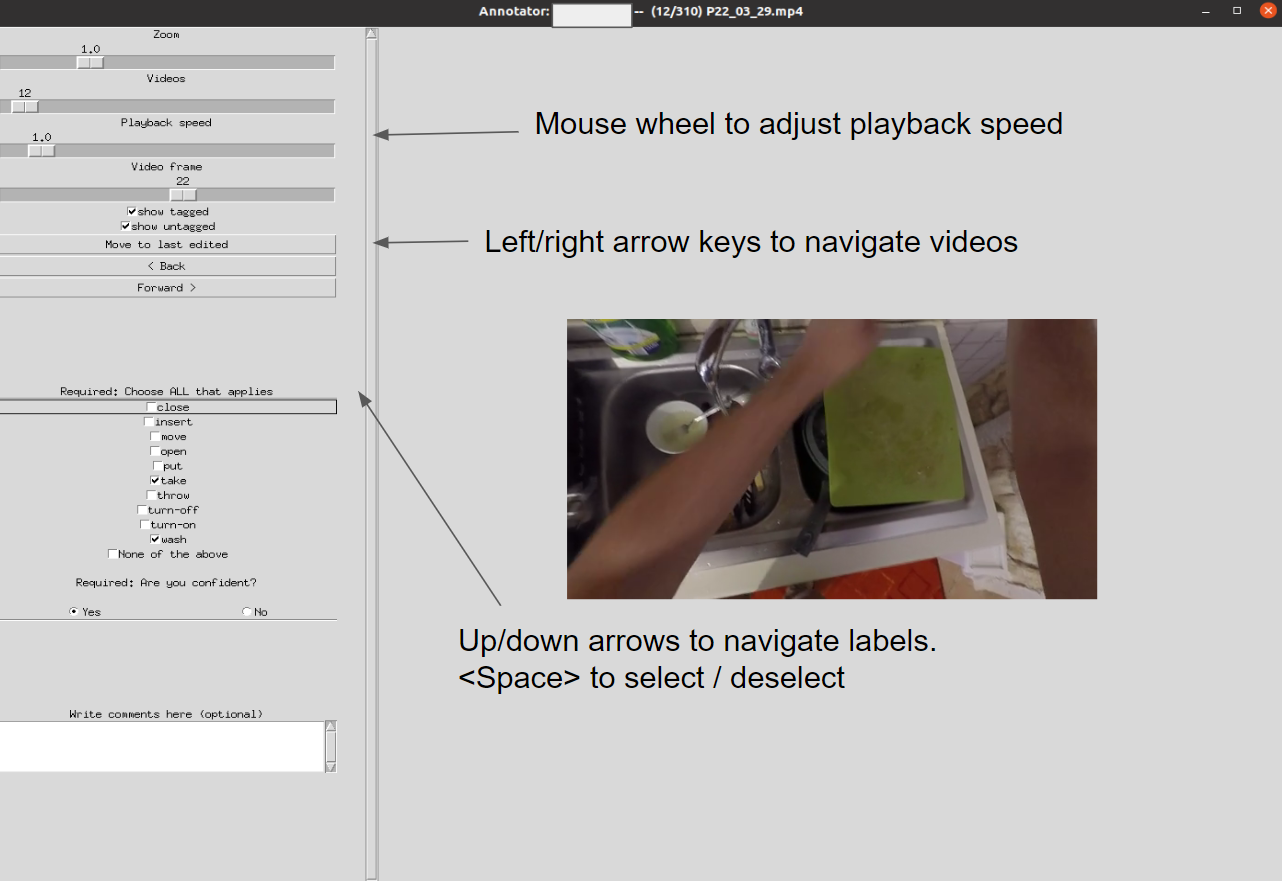}
    \vspace{-5pt}
    \caption{Our annotation interface.}

    \label{fig:annotation_tool}
\end{figure}

Here we describe the annotation protocol we followed to compile our EPIC-Kitchens-100-SPMV dataset. 
\Cref{fig:multilabel_cooccurrence} shows the co-occurrence frequency of head classes. \Cref{fig:annotation_tool} depicts our annotation interface.

\paragraph{Annotating difficult samples.}
We first trained a TSM~\cite{tsm} RGB model on EPIC-Kitchens-100~\cite{damen2020rescaling}. This model achieved 60.9\% verb accuracy on the original validation set. We then annotated %
failure cases where the top-1 prediction was incorrect, which amounted to 39.1\% of EPIC-Kitchens-100's validation set (3,782 videos).

\paragraph{Verb candidates.}
Annotators were presented with a video containing an action and were asked to choose any verb they saw fit to describe the action. Annotators could also choose no labels in case the video was noisy and did not show any action. Annotators could not see the original ground truth and could only choose from a limited list of verbs.
We exploited the TSM model predictions to compile a list of verb candidates for each video. 
Specifically, we observed that in many cases the model was predicting reasonable verbs within its top predictions. 
To facilitate labelling we thus %
show annotators only the 10 verbs corresponding to the top 10 predictions of the model for a given video.
We also included the ground truth single-verb label among the verb list (without telling annotators that this was the ground truth). 
This was to analyse the performance of the annotators: if they did not choose the ground truth verb in most cases then they likely did not pay enough attention to the video. 
Regardless of the chosen verbs we always included the original ground truth in the final multi-labels. 

\paragraph{Multiple annotators.}
For robustness each video was annotated by at least three different annotators. %
In some cases annotators did not meet our quality requirements, \eg they chose too few verbs on average per video or failed to choose the ground truth most of the time.
In these cases we asked other annotators to label the same videos. In total we employed 26 annotators. People were students mostly from our computer science department.

\paragraph{Filtering noisy samples.}
For each video annotators were asked whether they were confident about their verb selection or not. We discarded samples where more than half annotators did not feel confident. Many of the discarded videos include actions happening outside the viewpoint, segments with bad temporal boundaries or where no action was happening.
Altogether, we filtered out 289 videos from the initial 3,782 samples, compiling a set of 3,493 clean annotations.

\paragraph{Finalising labels.}
For a given video we keep labels that were chosen by at least half of the annotators. For example, for a video annotated by three people a verb would have to be chosen at least twice to be kept in the video multi-verb annotations. We also included the original single-label ground truth regardless of the annotators' choice.

\paragraph{Held-out validation set.}
Our 3,493 multi-verb annotations constitute our test set. The remaining $9,668 - 3,493 = 6,175$ videos from EPIC-Kitchens-100's validation set constitute our own validation set, where videos are annotated with only one verb.

\paragraph{Fleiss' kappa agreement score.}
\cameraready{We computed Fleiss' kappa to measure agreement. We report an average $\kappa$ of 0.52, which is typically translated as moderate agreement. Labelling actions is highly subjective, thus we believe annotators showed a satisfactory agreement.}

\section{SPML Losses}
In this section we formally define all the SMPL losses introduced in the paper.
For convenience, $f_i^{(c)} = f^{(c)}(\mathbf{x}_i)$ denotes the model prediction confidence for class $c$.
Firstly, Weak Assume Negative (WAN) loss~\cite{cole2021multi} has a negative balancing weight $\frac{1}{C}$ from the Assume Negative (AN) loss:
\begin{align}
    \mathcal{L_{\text{WAN}}}(\mathbf{x}_i, y_i) = -\frac{1}{C} \sum_{c=1}^{C}{\left[ \mathbb{1}_{[y_i=c]}\log{(f^{(c)}(\mathbf{x}_i))} + \mathbb{1}_{[y_i\neq c]} \frac{1}{C-1} \log{(1-f^{(c)}(\mathbf{x}_i))}\right]} 
\end{align}

\noindent Label Smoothing (LS) loss~\cite{szegedy2016rethinking} is defined as:

\begin{align}
    \mathcal{L_{\text{LS}}}(\mathbf{x}_i, y_i) = -\frac{1}{C} \sum_{c=1}^{C}{\left[ \mathbb{1}_{[y_i=c]}^\frac{\epsilon}{2}\log{(f^{(c)}(\mathbf{x}_i))} + \mathbb{1}_{[y_i\neq c]}^\frac{\epsilon}{2} \log{(1-f^{(c)}(\mathbf{x}_i))}\right]} 
\end{align}

\noindent where $\mathbb{1}_{[Q]}^\alpha = (1-\alpha)\mathbb{1}_{[Q]} + \alpha\mathbb{1}_{[\neg Q]}$ for any logical preposition $Q$. The label smoothing only for assumed negatives (N-LS)~\cite{zhou2022acknowledging} would be the same only for the assumed negatives:

\begin{align}
\begin{split}
    \mathcal{L_{\text{N-LS}}}(\mathbf{x}_i, y_i) = -\frac{1}{C} \sum_{c=1}^{C} &\bigg[ \mathbb{1}_{[y_i=c]}\log{(f^{(c)}(\mathbf{x}_i))} \\
    &+ \mathbb{1}_{[y_i\neq c]} \left[(1-\epsilon)\log{(1-f^{(c)}(\mathbf{x}_i))} + \epsilon \log{(f^{(c)}(\mathbf{x}_i))}\right]\bigg] 
\end{split}
\end{align}

\noindent We used $\epsilon = 0.1$ throughout the paper. Focal loss~\cite{lin2017focal} has a balancing parameter $\alpha$ and a focussing parameter $\gamma$.

\begin{align}
\begin{split}
    \mathcal{L_{\text{Focal}}}(\mathbf{x}_i, y_i) = -\frac{1}{C} \sum_{c=1}^{C} &\bigg[ \mathbb{1}_{[y_i=c]} \left[ \alpha (1-f^{(c)}(\mathbf{x}_i))^\gamma \log{(f^{(c)}(\mathbf{x}_i))}\right]\\
    &+ \mathbb{1}_{[y_i\neq c]} \left[ (1-\alpha) (f^{(c)}(\mathbf{x}_i))^\gamma \log(1-f^{(c)}(\mathbf{x}_i))\right]\bigg]
\end{split}
\end{align}

\noindent We set $\alpha = 0.25$ and $\gamma = 2$. Finally, the Entropy Maximisation loss~\cite{zhou2022acknowledging} is defined as follows:

\begin{align}
    \mathcal{L_{\text{EM}}}(\mathbf{x}_i, y_i) &= -\frac{1}{C} \sum_{c=1}^{C}{\left[ \mathbb{1}_{[y_i=c]}\log{(f^{(c)}(\mathbf{x}_i))} + \mathbb{1}_{[y_i\neq c]} \alpha H(f^{(c)}(\mathbf{x}_i))\right]}\\
    H(f^{(c)}(\mathbf{x}_i)) &= -\left[f^{(c)}(\mathbf{x}_i)\log{\left(f^{(c)}(\mathbf{x}_i)\right)} + (1-f^{(c)}(\mathbf{x}_i))\log{\left(1-f^{(c)}(\mathbf{x}_i)\right)}\right]
\end{align}

\section{Definitions of IOU Accuracy and F\textsubscript{1}}
Let $N$ be the number of videos in the test set, $\mathcal{Y}_i$ be the set of ground truth, and $\hat{\mathcal{Y}}_i$ be the set of predicted classes with over 50\% prediction confidence. The IOU accuracy and F\textsubscript{1}-Measure are defined as follow:
\paragraph{IOU accuracy:}

\begin{align}
    A_{\text{IOU}} = \frac{1}{N} \sum_{i=1}^N{\frac{|\mathcal{Y}_i \cap \hat{\mathcal{Y}}_i|}{|\mathcal{Y}_i \cup \hat{\mathcal{Y}}_i|}}
\end{align}

\paragraph{F\textsubscript{1}-Measure:}

\begin{align}
    F_1 = \frac{1}{N} \sum_{i=1}^N{\frac{2|\mathcal{Y}_i \cap \hat{\mathcal{Y}}_i|}{|\mathcal{Y}_i| + |\hat{\mathcal{Y}}_i|}}
\end{align}

\section{Impact of $K$ and $\tau$}

There are two hyperparameters involved in our methods: the number of neighbours ($K$) and the pseudo-label threshold ($\tau$). \Cref{tab:param_search} shows the impact of the different choices of $K$ and $\tau$. 
Results improve with a larger number of neighbours and a smaller $\tau$, which entail a larger set of pseudo-labels. This shows the benefits of using more pseudo-labels to better alleviate semantic ambiguity.

\begin{table}[t]

\centering

\begin{subtable}[t]{0.4\textwidth}
    \centering
    \caption{Mask BCE loss}
    \label{tab:somethingv1}
\resizebox{0.95\linewidth}{!}{
\begin{tabular}{lccccc}
\hline
$\tau$/K & 3 & 5 & 10 & 15 & 20 \\ \hline
0.1      & 27.2  & 30.5  & 31.0  & 36.4  & 36.2  \\
0.2      & 27.2  & 23.8  & 26.3  & 27.5  & 28.6  \\
0.3      & 27.2  & 23.8  & 23.9  & 25.8  & 25.1  \\ \hline
\end{tabular}}%
    \vspace{7pt}
\end{subtable}%
\qquad
\begin{subtable}[t]{0.4\textwidth}
    \centering
    \caption{P+S BCE loss}
    \label{tab:somethingv2}
\resizebox{0.95\linewidth}{!}{
\begin{tabular}{lccccc}
\hline
$\tau$/K & 3 & 5 & 10 & 15 & 20 \\ \hline
0.1      & 37.9  & 42.7  & 41.6  & 44.5  & 43.7  \\
0.2      & 37.9  & 30.5  & 34.3  & 34.7  & 37.6  \\
0.3      & 37.9  & 30.5  & 29.7  & 33.5  & 31.9  \\ \hline
\end{tabular}}
    
\end{subtable}%

\caption{F\textsubscript{1}-Measure for our Mask and P+S BCE loss on EPIC-Kitchens-100-SPMV with different values of $\tau$ and K.
}
\label{tab:param_search}
\vspace{-15pt}
\end{table}

\section{Qualitative results}

\begin{figure}[t]
    \centering
    \includegraphics[width=\textwidth]{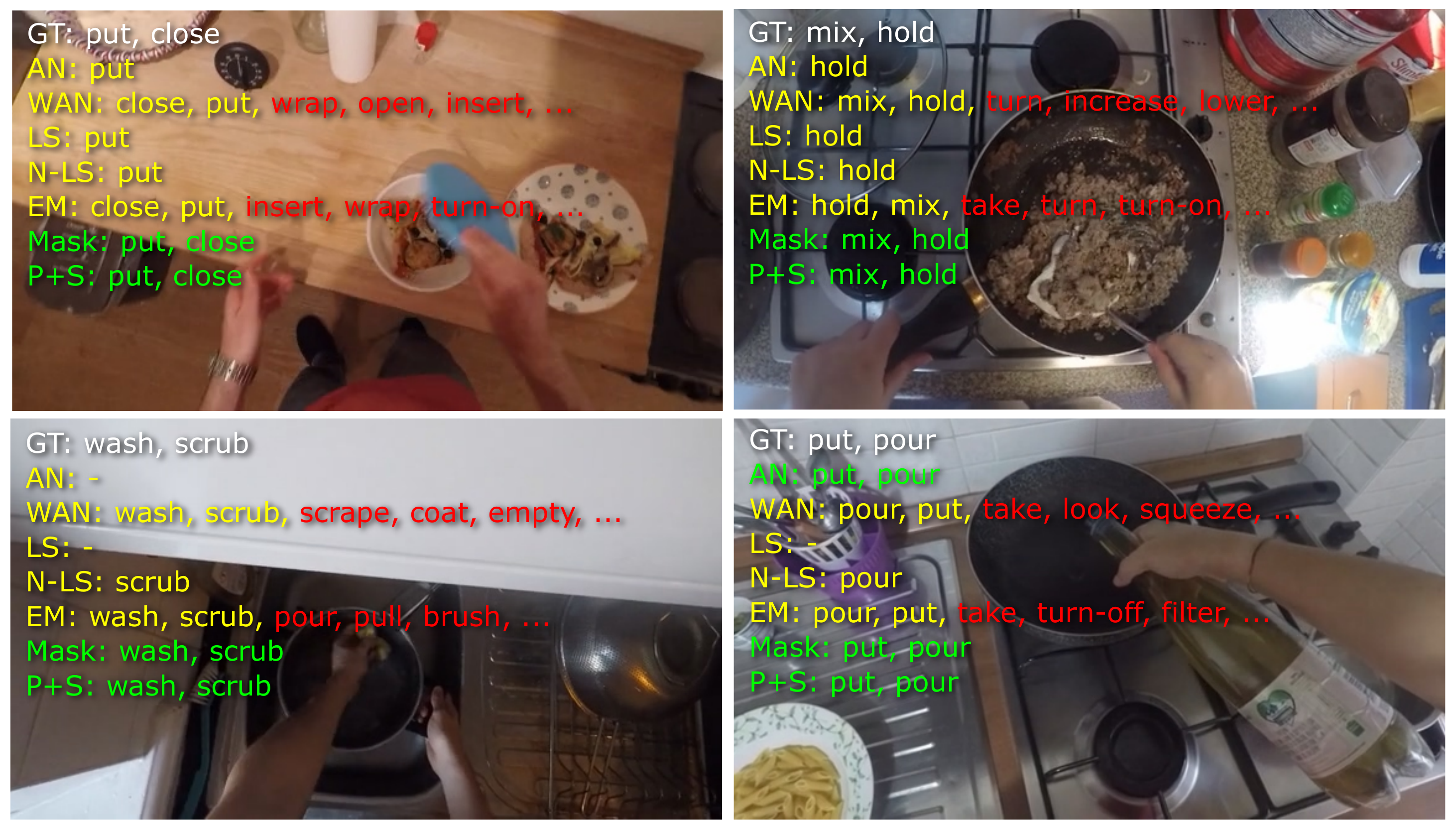}
    \vspace{-5pt}
    \caption{Qualitative examples from our methods and the other SPML baselines. White indicates the ground truth, yellow denotes a partial match, while red and green denote incorrect and successful total matches, respectively. We cap the model predictions shown here to five. A dash ``--'' denotes the case where the are no predictions due to the model having low confidence for all classes.} %

    \label{fig:qualitative}
\end{figure}

Figure~\ref{fig:qualitative} illustrates a few qualitative examples where we compare the predictions obtained with our methods and the other SPML baselines. 
Verbs shown here were selected by thresholding the model confidence (sigmoid) at 50\%. A ``-'' indicates no predictions, which happens when all classes have low confidence.
We note that WAN and EM tend to over-predict numerous verbs, whereas AN, LS and N-LS tend to predict only one verb. Our methods instead correctly output relevant verbs without over-predicting a large number of labels.

\section{End-to-end experiments}

We additionally performed end-to-end training on the Confusing-HMDB-102 dataset, using optical flow images. Optical flow was extracted using the TV-L1~\cite{zach2007duality} algorithm. In this case the training set features and the pseudo labels were updated every 5 epochs. \Cref{tab:hmdb} reports results obtained with this setting. 
Performance with end-to-end training naturally improves compared to results reported in the main paper, where pre-extracted features were used throughout training.
We observe that our methods consistently outperform all baselines. 

\begin{table}[t]

\centering

\begin{tabular}{|l|c|c|c|c|c|}
\hline
Loss & Top-set ML & Top-1 ML & IOU Acc. & F\textsubscript{1} & mAP\\
\hline
\hline
AN & \roundnumpm{46.72}{1.49} & \roundnumpm{49.52}{1.68} &
\roundnumpm{11.50}{0.81} &
\roundnumpm{13.68}{1.06} &
\roundnumpm{47.78}{1.24}\\
WAN & \roundnumpm{38.23}{2.37} & \roundnumpm{40.94}{2.66} &
\roundnumpm{13.73}{0.87} &
\roundnumpm{22.99}{1.28} &
\roundnumpm{37.39}{2.16}\\

LS & \roundnumpm{47.77}{0.81} &
\roundnumpm{51.02}{0.49} &
\roundnumpm{9.46}{1.19} &
\roundnumpm{11.43}{1.46} &
\roundnumpm{49.78}{0.75}\\

N-LS & \roundnumpm{48.57}{1.34} &
\roundnumpm{51.07}{1.16} &
\roundnumpm{11.97}{2.21} &
\roundnumpm{14.19}{2.84} &
\roundnumpm{49.95}{2.12}\\

EM & \roundnumpm{47.69}{1.72} &
\roundnumpm{48.54}{1.73} &
\roundnumpm{3.36}{0.05} &
\roundnumpm{6.49}{0.10} &
\roundnumpm{49.58}{2.75}\\
\hline

Mask &
\underline{\roundnumpm{50.41}{0.11}} &
\textbf{\roundnumpm{53.20}{0.23}} &
\underline{\roundnumpm{25.57}{1.77}} &
\underline{\roundnumpm{27.86}{1.95}} &
\underline{\roundnumpm{51.89}{1.21}}\\

P+S &
\textbf{\roundnumpm{50.45}{0.31}} &
\underline{\roundnumpm{51.79}{0.25}} &
\textbf{\roundnumpm{30.88}{2.34}} &
\textbf{\roundnumpm{33.38}{2.59}} &
\textbf{\roundnumpm{52.35}{1.27}}\\

\hline

Mask\ssymbol{1} &
\roundnumpm{53.40}{0.90} &
\roundnumpm{54.70}{1.48} &
\roundnumpm{32.53}{0.81} &
\roundnumpm{34.32}{0.89} &
\roundnumpm{55.42}{1.13}\\

P+S\ssymbol{1} &
\roundnumpm{57.15}{0.63} &
\roundnumpm{57.25}{0.79} &
\roundnumpm{41.71}{2.07} &
\roundnumpm{42.71}{1.91} &
\roundnumpm{59.88}{1.26}\\

\hline
\end{tabular}%
\vspace{7pt}
\qquad

\caption{Results obtained training the model end-to-end with optical flow images. Results in percent (\%) $\pm$ standard deviation on the three splits of Confusing-HMDB-102. Losses marked with \nsymbol{1} assume the case when the pseudo label search is ideal, and gets to use actual ground truth labels during training. Mask and P+S (without \nsymbol{1}) are trained with $K=10$ and $\tau=0.2$.
}

\label{tab:hmdb}
\end{table}

\end{document}